# NP-Free: A Real-time Normalization-free and Parameter-turning-free Representation Approach for Open-ended Time Series

Ming-Chang Lee[1], Jia-Chun Lin[2], and Volker Stolz[3]

[1,3]Department of Computer science, Electrical engineering and Mathematical sciences, Høgskulen på Vestlandet (HVL), Bergen, Norway

[2]Department of Information Security and Communication Technology, Norwegian University of Science and Technology, Gjøvik, Norway

[1] mingchang1109@gmail.com; ming-chang.lee@hvl.no
[2] jia-chun.lin@ntnu.no
[3] volker.stolz@hvl.no

# NP-Free: A Real-Time Normalization-free and Parameter-tuning-free Representation Approach for Open-ended Time Series

Ming-Chang Lee
Dept. of Computer science, Electrical engineering and Mathematical sciences
Høgskulen på Vestlandet (*HVL*)
Bergen, Norway
mingchang1109@gmail.com; ming-chang.lee@hvl.no

Jia-Chun Lin
Dept. of Information Security and Communication Technology
Norwegian University of Science and Technology (*NTNU*)
Gjøvik, Noway
jia-chun.lin@ntnu.no

Volker Stolz
Dept. of Computer science, Electrical engineering and Mathematical sciences
Høgskulen på Vestlandet (*HVL*)
Bergen, Norway
volker.stolz@hvl.no

*Abstract*— As more connected devices are implemented in a cyber-physical world and data is expected to be collected and processed in real time, the ability to handle time series data has become increasingly significant. To help analyze time series in data mining applications, many time series representation approaches have been proposed to convert a raw time series into another series for representing the original time series. However, existing approaches are not designed for open-ended time series (which is a sequence of data points being continuously collected at a fixed interval without any length limit) because these approaches need to know the total length of the target time series in advance and pre-process the entire time series using normalization methods. Furthermore, many representation approaches require users to configure and tune some parameters beforehand in order to achieve satisfactory representation results. In this paper, we propose NP-Free, a real-time <u>N</u>ormalization-free and <u>P</u>arameter-tuning-<u>free</u> representation approach for open-ended time series. Without needing to use any normalization method or tune any parameter, NP-Free can generate a representation for a raw time series on the fly by converting each data point of the time series into a root-mean-square error (RMSE) value based on Long Short-Term Memory (LSTM) and a Look-Back and Predict-Forward strategy. To demonstrate the capability of NP-Free in representing time series, we conducted several experiments based on real-world open-source time series datasets. We also evaluated the time consumption of NP-Free in generating representations.

## I. Introduction

In recent years, there has been an increasing need for time series analysis such as clustering, classification, anomaly detection, forecasting, indexing, etc. [2] [5] [6] [18] [25] [26] due to the explosion of the Internet of Things in a cyber-physical world. Large volumes of time series data are continuously measured and collected from connected devices and sensors. Analyzing raw time series can be challenging and undesirable due to high computation cost and memory requirement [13]. One solution to achieve effective and efficient time series analysis is via high-level representation approaches, which aim to extract features of time series or reduce the dimensionality of the time series while preserving the fundamental characteristics of the time series [16].

A number of time series representation approaches have been proposed in the last few years. However, many of them (such as Symbolic Aggregate approXimation [8], Piecewise Aggregate Approxmiation [12], and the clipped representation approach [19]) work only on fixed-length time series, rather than on open-ended time series. This is something to do with their underlying designs. Before generating a representation for a time series, these approaches need to preprocess the time series using z-normalization (also known as z-score normalization), which is a commonly used normalization approach for time series [14]. However, z-normalization might cause two distinct time series to become indistinguishable [7], therefore misleading the representation approaches and negatively affecting subsequent data mining tasks. Another shortcoming with most representation approaches is that users require to configure and tune certain parameters (e.g., the length of the time series, the size of sliding window, or the size of Alphabet, etc. [8]) in advance. Without a proper value to each parameter, these approaches may generate improper representations for time series, consequently affecting subsequent data mining tasks.

To address aforementioned issues, this paper introduces NP-Free, which is a real-time <u>N</u>ormalization-free and <u>P</u>arameter-tuning-<u>free</u> representation approach for open-ended time series. Inspired by one of the state-of-the-art real-time time series anomaly detection approaches called RePAD [2], NP-Free adopts Long Short-Term Memory (LSTM) and the Look-Back and Predict-Forward strategy used by RePAD to generate representations for time series. Note that NP-Free does not need to preprocess time series using any normalization approach or require users to tune any parameter. In fact, NP-Free can directly generate a representation for a raw time series on the fly by transforming the time series into a root-mean-square error (RMSE) series in real time (i.e., instantaneously).

More specifically, NP-Free keeps predicting the next data point in the target time series based on three historical data points and keeps calculating the corresponding RMSE value between the observed and predicted data points. By doing so,

NP-Free can convert the target time series into a RMSE series. To adapt to any pattern change in the time series, NP-Free also retrains a new LSTM model as needed based on the self-adaptive detection threshold introduced in [2].

NP-Free can be applied to any open-ended univariate time series with any pattern (i.e., recurrent and non-recurrent) since NP-Free does not require the knowledge of the total length of the target time series in advance. To demonstrate the capability of NP-Free in representing time series, we performed a series of experiments based on real-world open-source time series datasets from both Numenta Anomaly Benchmark (NAB) [3] and Melbourne Pedestrian Foot Traffic [33]. The experiment results demonstrate that the representations generated by NP-Free can retain the fundamental structural properties of the corresponding time series and can therefore represent the time series.

The contributions of NP-Free are as follows:

1. Normalization-free: NP-Free can directly generate a representation for any raw univariate time series without requiring to process the time series using any normalization approach beforehand.
2. Parameter-tuning-free: NP-Free does not require users to tune or configure any hyperparameters or parameters in advance. All hyperparameters and parameters in NP-Free are already pre-fixed, and they were used or suggested by [2] and [4].
3. No human intervention: NP-Free can automatically train or retrain its LSTM model when it is needed. Users are not required to do anything to use NP-Free.
4. Real-time and on the fly: NP-Free can transform an open-ended time series into a RMSE series on the fly instantaneously. Therefore, it can be employed by real-time time series analysis applications (such as clustering, classification, etc.) to preprocess time series.
5. Lightweight: NP-Free is extremely lightweight because it employs a simple LSTM network (one hidden layer with ten hidden units), always uses three data points for LSTM model training, and only retrains a new LSTM model when necessary.

The rest of the paper is organized as follows: Section II presents background and related work. In Section III, we introduce the details of NP-Free. Section IV presents and discusses the experiments and the corresponding results. Section V concludes this paper and outlines future work. In the Appendix, we demonstrate that z-normalization might mislead time series representation approaches and negatively impact subsequent time series analysis tasks.

## II. BACKGROUND AND RELATED WORK

This section introduces z-normalization, time series representation approaches related to this paper, and RePAD [2]

### A. Z-normalization

Z-normalization (also known as z-score normalization) refers to the process of normalizing every value in a dataset such that the mean of all of the values is approximately 0 while the standard deviation is in a range close to 1 [20]. The formula of z-normalization is shown below.

$$\bar{x}_i = \frac{x_i - \mu}{\sigma} \quad (1)$$

where $x_i$ is the $i$-th data point in the time series, $\mu$ is the mean of all data points in the time series, $\sigma$ is the standard deviation of all the data points, and $\bar{x}_i$ is normalized data point derived from the equation.

Z-normalization is considered an essential preprocessing step which allows time series representation approaches to focus on the structural similarities/dissimilarities of time series rather than on the original data point values [20]. However, z-normalization has some shortcomings. When it encounters a flat time series, any fluctuations (e.g., noises) will be enhanced, resulting in an negative impact to a data mining technique, such as Matrix Profile [17]. According to [7], z-normalization might also destroy potentially relevant properties to distinguish time series since the time series are shifted to have zero mean. In other words, two distinct time series might become indistinguishable after z-normalization. Similarly, according to [14] [20] [21], z-normalization might not produce normalized data with the exact same scale, causing a negative impact to subsequent data mining tasks. We will also show this in the Appendix.

### B. Time series representation approaches

As mentioned earlier, time series representations are often treated as the first component of time series analysis [16]. According to [13], time series representation approaches can be classified into two categories: data adaptive and non-data adaptive. Data adaptive representation methods, including Symbolic Aggregate approXimation (SAX) [8] and the clipped representation approach [19], try to minimize the global reconstruction error using arbitrary length segments [16]. According to [16], this type of methods can better approximate each time series in datasets, but the comparison of several time series is difficult. In contrast, non-data adaptive approaches, including Piecewise Aggregate Approxmiation (PAA) [12] and Discrete Fourier Transformation (DFT) [32], considered local properties of time series and constructed an approximate representation accordingly [28]. Hence, non-data adaptive approaches are suitable for time-series with equal-length segmentation, and it is straightforward to compare the representations of several time series [16].

Despite the above classification, existing representation approaches are designed to generate representations for fixed-length time series because they require to normalize time series using z-normalization before generating representations for time series. Example approaches include SAX [8], PAA [12], Symbolic Aggregate approXimation and Vector Space Model (SAX-VSM) [15], the clipped representation approach [19], etc. Furthermore, they also require users to set some parameters in advance. Without a proper value to each parameter, these approaches might not be able to generate proper representations for time series. For instance, SAX requires users to decide a desired word size and choose an alphabet size.

Different from these time series representation approaches, NP-Free is a novel representation approach since it is normalization-free and parameter-tuning-free. It does not suffer from the issues caused by z-normalization or require users to tune any hyperparameter or parameter. NP-Free is able to generate representations for open-ended time series on the fly. To our best knowledge, NP-Free is the first real-time representation approach for open-ended time series.

*C. RePAD: The building block of NP-Free*

Recall that NP-Free is inspired by RePAD [2], which is a lightweight real-time anomaly detection approach for univariate time series. In order to help readers easily understand NP-Free and how NP-Free is different from RePAD, we introduce RePAD in this section.

RePAD utilizes short-term historical data points to predict and determine whether each upcoming data point is anomalous or not based on LSTM, the Look-Back and Predict-Forward strategy [36], and a self-adaptive detection threshold. LSTM is an neural network designed to learn long short-term dependencies and model temporal sequences [1]. Due to the small size of training data, RePAD employs a simple LSTM network (one hidden layer with only ten hidden units) to keep itself as lightweight as possible. Furthermore, to adapt to data pattern changes in time series, RePAD dynamically adjusts its threshold to detect anomalies and performs LSTM model retraining only when it is necessary. This features make RePAD adaptive and able to offer real-time anomaly detection.

Fig. 1 shows the algorithm of RePAD where $b$ is the Look-Back parameter. In other words, RePAD always predicts each upcoming data point based on $b$ historical data points. Let $t$ be the current time point and $t$ starts from 0, which is the time point when RePAD is launched. Let $v_t$ be the real data point collected at time point $t$, and $\widehat{v_t}$ be the data point predicted by RePAD at time point $t$.

Suppose $b$ is set to 3. At time points 0 and 1, RePAD has to collect data points $v_0$ and $v_1$, respectively. When $t$ equals 2, RePAD trains an LSTM model called $M$ by taking the three observed data points $v_0$, $v_1$, and $v_2$ as the training data. Then the model $M$ is immediately used to predict the value of the next data point, denoted by $\widehat{v_3}$. When $t$ equals 3, RePAD uses data points $v_1$, $v_2$, and $v_3$ to train an LSTM model, and uses it to predict the next data point, denoted by $\widehat{v_4}$. Similarly, when $t$ equals 4, RePAD uses data points $v_2$, $v_3$, and $v_4$ to predict $\widehat{v_5}$.

When $t$ equals 5, RePAD calculates its prediction error at the moment, denoted by $AARE_5$, based on a well-known prediction accuracy metric called Average Absolute Relative Error (AARE for short) [35] as shown in Equation 2.

$$AARE_t = \frac{1}{b} \cdot \sum_{y=t-b+1}^{t} \frac{|v_y - \widehat{v_y}|}{v_y}, t \geq 2b - 1 \quad (2)$$

A low AARE value indicates that the predicted values are close to the observed values. After deriving $AARE_5$, RePAD uses data points $v_3$, $v_4$, and $v_5$ to predict $\widehat{v_6}$ (see lines 9 and 10 of Fig. 1). When $t$ equals 6, RePAD repeats the same procedure to calculate $AARE_6$ and predict $\widehat{v_7}$.

When $t$ equals 7, RePAD calculates $AARE_7$. Also, RePAD can officially detect anomalies since it has sufficient AARE values (i.e., $AARE_5$, $AARE_6$, and $AARE_7$) to calculate a detection threshold called $thd$. In other words, in order to get the detection function started, RePAD requires a preparation period of 7 time points. Equation 3 shows how $thd$ is calculated based on the Three-Sigma Rule [34].

$$thd = \mu_{AARE} + 3 \cdot \sigma, t \geq 2b + 1 \quad (3)$$

where $\mu_{AARE}$ is the average AARE as shown in Equation 4, and $\sigma$ is the standard deviation as shown in Equation 5.

$$\mu_{AARE} = \frac{1}{t - b - 1} \cdot \sum_{x=2b-1}^{t} AARE_x \quad (4)$$

$$\sigma = \sqrt{\frac{\sum_{x=2b-1}^{t}(AARE_x - \mu_{AARE})^2}{t - b - 1}} \quad (5)$$

As shown in line 15 of Fig. 1, if $AARE_7$ is lower than or equal to $thd$, data point $\widehat{v_7}$ is not considered anomalous, and the current LSTM model will be kept for future prediction (i.e., LSTM retraining is not required). However, if $AARE_7$ is higher than $thd$ (see line 16), it might indicate that the data pattern of the time series have changed or that anomalies might have happened. In this case, RePAD tries to adapt to the change by retraining an new LSTM model with data points $v_4$, $v_5$, and $v_6$ to re-predict $\widehat{v_7}$ and re-calculate both $AARE_7$ and $thd$.

```
RePAD algorithm
Input: Data points in a time series
Output: Anomaly notifications
Procedure:
1:   Let t be the current time point and t starts from 0; Let flag be True;
2:   While time has advanced {
3:     Collect data point v_t;
4:     if t ≥ b − 1 and t < 2b − 1 {  // i.e., 2 ≤ t < 5, if b = 3
5:       Train an LSTM model by taking [v_{t−b+1}, v_{t−b+2} …, v_t] as the training data;
6:       Let M be the resulting LSTM model and use M to predict v_{t+1};}
7:     else if t ≥ 2b − 1 and t < 2b + 1 {  //i.e., 5 ≤ t < 7, if b = 3
8:       Calculate AARE_t based on Equation 2;
9:       Train an LSTM model by taking [v_{t−b+1}, v_{t−b+2} …, v_t] as the training data;
10:      Let M be the resulting LSTM model and use M to predict v_{t+1};}
11:    else if t ≥ 2b + 1 and flag==True {  //i.e., t ≥ 7 if b = 3
12:      if t ≠ 7 { Use M to predict v_t;}
13:      Calculate AARE_t based on Equation 2;
14:      Calculate thd based on Equation 3;
15:      if AARE_t ≤ thd { v_t is not considered as an anomaly;}
16:      else{
17:        Train an LSTM model with [v_{t−b}, v_{t−b+1}, …, v_{t−1}];
18:        Use the newly trained LSTM model to predict v_t;
19:        Calculate AARE_t using Equation 2;
20:        Calculate thd based on Equation 3;
21:        if AARE_t ≤ thd { v_t is not considered as an anomaly;}
22:        else {
23:          v_t is reported as an anomaly immediately;
24:          Let flag be False;}}}
25:    else if t ≥ 2b + 1 and flag==False {
26:      Train an LSTM model with [v_{t−b}, v_{t−b+1}, …, v_{t−1}];
27:      Use the newly trained LSTM model to predict v_t;
28:      Calculate AARE_t based on Equation 2;
29:      Calculate thd based on Equation 3;
30:      if AARE_t ≤ thd{
31:        v_t is not considered as an anomaly;
32:        Replace M with the new LSTM model from line 26;
33:        Let flag be True;}
34:      else {
35:        v_t is reported as an anomaly immediately; Let flag be False;}}}
```

Fig. 1. The algorithm of RePAD [2].

If the re-calculated $AARE_7$ is lower than or equal to $thd$ (see line 21), $v_7$ is not considered anomalous. Otherwise, $v_7$ is considered as anomalous because the new trained LSTM model is still unable to accurately predict $v_7$. In this case, RePAD will set its $flag$ to "False" (see line 24), enabling a new LSTM model to be trained at the next time point instead of using the same model. The above detection process will repeat over and over again as time advances.

### III. METHODOLOGY OF NP-FREE

With the knowledge about RePAD, now we can introduce how NP-Free works. As mentioned earlier, NP-Free aims to convert an open-ended univariate time series into a RMSE series in real time for representing the time series. To achieve the above goal, NP-Free adopts many strategies used by RePAD, including the simple LSTM network structure (i.e., one hidden layer with ten hidden units), the Look-Back and Predict-Forward strategy, and the main idea of the detection threshold. In addition, NP-Free follows the suggestion from [4] to set the Look-Back parameter to 3. In other words, NP-Free always uses three historical data points to predict each upcoming data point.

However, different from RePAD, NP-Free is not proposed to detect anomalies. Instead, it is used to predict every upcoming data point in the target time series and calculate the corresponding RMSE value. To make sure the deterministic property of NP-Free (i.e., the same RMSE series will always be generated given the same time series), NP-Free removes all randomness by fixing all LSTM hyperparameter setting (including the number of hidden layers, the number of hidden units, learning rate, the number of epochs, etc.) and following the hyperparameter setting used by RePAD [2].

Fig. 2 illustrates the algorithm of NP-Free where $T$ denotes the current time point and $T$ starts from 0, which is the time point when NP-Free is launched. Let $d_T$ be the real data point collected at time point $T$, and $\widehat{d_T}$ be the data point predicted by NP-Free at time point $T$. It is clear that the algorithm is similar to that of RePAD as shown in Fig. 1, so we will not repeat the same introduction. The main differences between NP-Free and RePAD are four folds: First, NP-Free always uses three data points to predict each upcoming data point. In other words, it removes the Look-Back parameter by fixing the value of it to three, followed by the suggestion made by [4].

The second difference is that NP-Free uses RMSE to measure its prediction error instead of using AARE. Equation 6 shows how $RMSE_T$ (i.e., the RMSE at time point $T$) is calculated.

$$RMSE_T = \sqrt{\frac{\sum_{z=T-2}^{T}(d_z - \widehat{d_T})^2}{3}}, T \geq 5 \quad (6)$$

It is clear that given any two time series (says $A$ and $B$), RMSE is to evaluate the absolute error between $A$ and $B$, meaning that the error would be always identical no matter $A$ is the observed time series or the predicted one. On the contrary, AARE is to evaluate the relative error. Hence, the error relative to $A$ might not be the same as the error relative to $B$. Due to the above reason, NP-Free chose RMSE to help achieve the above-mentioned deterministic property.

The third difference is that NP-Free keeps calculating its threshold, denoted by $thd_{RMSE}$ (see Equation 7), based on a fixed number of RMSE values, rather than based on all historical AARE values.

```
NP-Free algorithm
Input: Data points in the target time series
Output: A RMSE series
Procedure:
1:   Let T be the current time point and T starts from 0; Let Flag be True;
2:   While time has advanced {
3:     Collect data point d_T;
4:     if T ≥ 2 and T < 5 {
5:       Train an LSTM model by taking d_{T-2}, d_{T-1}, and d_T as the training data;
6:       Let m be the resulting LSTM model and use m to predict d̂_{T+1};}
7:     else if T ≥ 5 and T < 7 {
8:       Calculate RMSE_T based on Equation 6 and output RMSE_T;
9:       Train an LSTM model by taking d_{T-2}, d_{T-1}, and d_T as the training data;
10:      Let m be the resulting LSTM model and use m to predict d̂_{T+1};}
11:    else if T ≥ 7 and Flag==True {
12:      if T ≠ 7 { Use m to predict d̂_T;}
13:      Calculate RMSE_T based on Equation 6;
14:      Calculate thd_{RMSE} based on Equation 7;
15:      if RMSE_T ≤ thd_{RMSE} { Output RMSE_T;}
16:      else {
17:        Train an LSTM model with d_{T-3}, d_{T-2}, and d_{T-1};
18:        Use the newly trained LSTM model to predict d̂_T;
19:        Calculate RMSE_T based on Equation 6;
20:        Calculate thd_{RMSE} based on Equation 7;
21:        if RMSE_T ≤ thd_{RMSE} { Output RMSE_T;}
22:        else { Output RMSE_T; Let Flag be False;}}}
23:    else if T ≥ 7 and Flag==False {
24:      Train an LSTM model with d_{T-3}, d_{T-2}, and d_{T-1};
25:      Use the newly trained LSTM model to predict d̂_T;
26:      Calculate RMSE_T based on Equation 6;
27:      Calculate thd_{RMSE} based on Equation 7;
28:      if RMSE_T ≤ thd_{RMSE} {
29:        Output RMSE_T;
30:        Replace m with the new LSTM model from line 24;
31:        Let Flag be True;}
32:      else { Output RMSE_T; Let Flag be False;}}}
```

Fig. 2. The algorithm of NP-Free.

$$thd_{RMSE} = M_{RMSE} + 3 \cdot \sigma \quad (7)$$

where $M_{RMSE}$ is the average RMSE at time point $T$, and it is derived as below.

$$M_{RMSE} = \begin{cases} \frac{1}{T-4} \cdot \sum_{x=5}^{T} RMSE_z, 7 \leq T < W+4 \\ \frac{1}{W} \cdot \sum_{x=T-W+1}^{T} RMSE_z, T \geq W+4 \end{cases} \quad (8)$$

where $W$ is the length of a sliding window to control how many previous RMSE values should be considered to derive the threshold. For instance, when $W$ is set to 100, NP-Free will always use the 100 most recently derived RMSE values to calculate $M_{RMSE}$ if the total number of historical RMSE values is sufficient. Since NP-Free is designed to generate representations for open-ended time series, it is important to include a limited sliding window so that the underlying system resources will not be exhausted by Equation 7.

Equation 9 shows how to calculate standard deviation $\sigma$ at time point $T$. Similarly, sliding window $W$ is also considered to derive $\sigma$. Whenever time point $T$ advances and it is greater than or equal to 7 (i.e., either line 11 of Fig. 2 or line 23 of Fig. 2 is evaluated to be true), NP-Free re-calculates $RMSE_T$ and $thd_{RMSE}$. If $RMSE_T$ is not greater than the threshold (see lines 15 and 28), $RMSE_T$ will be immediately outputted by NP-Free.

$$\sigma = \begin{cases} \sqrt{\dfrac{\sum_{x=5}^{T}(RMSE_z - M_{RMSE})^2}{T-4}}, 7 \leq T < W+4 \\ \sqrt{\dfrac{\sum_{x=T-W+1}^{T}(RMSE_z - M_{RMSE})^2}{W}}, T \geq W+4 \end{cases} \quad (9)$$

Otherwise, it might mean that the data pattern of the target time series has changed. In this case, NP-Free will try to adapt to the pattern change by retraining an new LSTM model to re-predict $\widehat{d_T}$ and re-calculate both $RMSE_T$ and $thd_{RMSE}$ either at current time point (i.e., lines 17 to 20) or at the next time point (i.e., lines 24 to 27) by setting its $Flag$ to be "False". If the re-calculated $RMSE_T$ is no larger than $thd_{RMSE}$, NP-Free immediately outputs $RMSE_T$. Otherwise, NP-Free outputs $RMSE_T$ and sets its $Flag$ to be "False", which will trigger LSTM model retraining at the next time point.

The last difference between NP-Free and RePAD is that NP-Free does not announce if a data point is anomalous or not because detecting anomalies is not its goal. By repeating the same process, an open-ended time series can be converted into a RMSE series on the fly in real time. In the next section, we will demonstrate the RMSE series generated by NP-Free can indeed represent the corresponding time series.

## IV. EXPERIMENT RESULTS

In this paper, we conducted two sets of experiments to demonstrate the capability of NP-Free in generating time series representations. The first set of the experiments is to show that NP-Free generates similar RMSE series when time series have a similar data pattern and shape. The second set of the experiments is to show that NP-Free does not generate similar RMSE series when time series have an opposite data pattern.

Note that we did not compare NP-Free with other representation approaches because none of them was designed to generate representations for open-ended time series in real time. In other words, we did not use $TLB$ (Tightness of Lower Bound) [13] to evaluate NP-Free even though $TLB$ is a common measure for comparing representation approaches. According to the equation of $TLB$, i.e., $TLB = LowerBoundDist(X,Y)/TrueEuclideanDist(X,Y)$, $X$ and $Y$ are two z-normalized time series. Since NP-Free works directly on raw time series, rather than on z-normalized time series, it is clear that $TLB$ cannot be applied to evaluate NP-Free. That is why we designed our own experiments to evaluate NP-Free.

TABLE I lists all hyperparameters and parameters used and pre-fixed by NP-Free. Note that all of them (except the sliding window) were used or suggested by [2] and [4]. Regarding the sliding window, we conducted an experiment to study how different sliding window sizes impact NP-Free. We chose 288, 576, 864, 1152, 1440, 1728, 2016, and 4032 to be the sliding window, but found that all of them had no significant impact to NP-Free in terms of representation generation and time consumption. Hence, we randomly chose one of them (i.e., 1440) to be the sliding window of NP-Free in all the experiments. In summary, users do not need to tune any hyperparameters or parameters for using NP-Free. That is why we said that NP-Free is parameter-tuning-free. As stated earlier, the main purpose for fixing all hyperparameters and parameters is to make sure that NP-Free can always produce the same RMSE series for the same time series (i.e., the deterministic property).

TABLE I.  THE HYPERPARAMETER AND PARAMETER SETTING USED BY NP-FREE.

| Hyperparameters and parameters | Value |
|---|---|
| The number of hidden layers | 1 |
| The number of hidden units | 10 |
| The number of epochs | 50 |
| Learning rate | 0.005 |
| Activation function | tanh |
| Random seed | 140 |
| The sliding window | 1440 |

In this paper, NP-Free was implemented in DeepLearning4J [37], and all the experiments were conducted on a MacBookPro laptop running MacOS Monterey 12.5.1 with Apple M1 Pro chip and 16GB Memory. The purpose of choosing this laptop is to show that NP-Free can be deployed on a commodity machine.

### A. Experiment Set 1

In experiment set 1, we chose three time series from an open-source repository called NAB [3] and one time series from Melbourne Pedestrian Foot Traffic [33]. TABLE II lists the details of the four time series. The first two time series are separately called rds-cpu-utilization-e47b3b (B3B for short) and ec2-cpu-utilization-825cc2 (CC2 for short), and both possess a non-recurrent data pattern. The last two time series are separately called art-daily-small-noise (ADSN for short) and Bourke Street Mall South (BSMS for short), and they possess a recurrent data pattern. The reason of choosing these time series is to show that NP-Free works on any time series regardless of their data patterns.

TABLE II.  FOUR OPEN-SOURCE TIME SERIES USED IN EXPERIMENT SET 1.

| Name | Time Period | Total number of data points | Time Interval |
|---|---|---|---|
| B3B | From 2014-04-10, 00:02 to 2014-04-23, 23:57 | 4032 | 5 min |
| CC2 | From 2014-04-10, 00:04 to 2014-04-24, 00:09 | 4032 | 5 min |
| ADSN | From 2014-04-01, 00:00 to 2014-04-14, 23:55 | 4032 | 5 min |
| BSMS | From 2016-01-01, 00:00 to 2016-06-30, 23:00 | 4368 | 1 hour |

To show that NP-Free always generates similar RMSE series for similar *non-recurrent* time series, we created 12 variants for time series B3B and named them B3B+100, B3B+200, ..., B3B+1000, B3B+1500, and B3B+2000 by increasing each data point value of the B3B time series by 100, 200, ..., 1000, 1500, and 2000, respectively. Due to page limit, we only show B3B and B3B+100 in Fig. 3. It is clear that these two time series have the exactly the same shape and data pattern but have different offsets in the Y axis. In fact, all the B3B variants have the same shape and data pattern as B3B.

Then we used NP-Free to generate a RMSE series for B3B and each B3B variant. Just for illustration, Fig. 4 depicts the RMSE series generated by NP-Free for B3B and B3B+100. Apparently, it is difficult to distinguish the two RMSE series with human eyes because they almost overlap with each other.

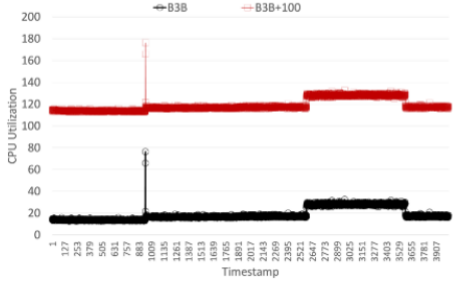

Fig. 3. All data points in B3B and B3B+100.

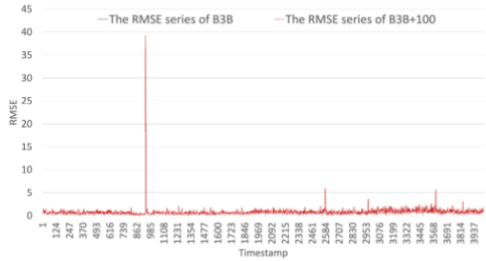

Fig. 4. The RMSE series generated by NP-Free for B3B and B3B+100.

Therefore, we utilized Euclidean Distance [11] to compare the two RMSE series. The formula of Euclidean Distance is shown below:

$$ED(R_i, R_j) = \sqrt{\sum_{z=1}^{L}(R_{i,z} - R_{j,z})^2} \quad (10)$$

where $R_i$ and $R_j$ are any two RMSE series ($i = 1, 2, \ldots, n$, and $j = 1, 2, \ldots, n$, but $i \neq j$), $R_{i,z}$ denotes the $z$-th data point in $R_i$, and $R_{j,z}$ denotes the $z$-th data point in $R_j$, where $z = 1, 2, \ldots, L$. Furthermore, we also calculated the Euclidean Distance between the RMSE series of B3B and the RMSE series of every other B3B variant separately. As listed in TABLE III, all Euclidean Distance values are very low, meaning that the RMSE series generated by NP-Free for all the B3B variants are individually similar to the RMSE series generated by NP-Free for B3B.

We repeated the same procedure to create 12 variants for CC2, and used NP-Free to generate a RMSE series for CC2 and each CC2 variant. TABLE IV shows that the Euclidean Distance between the RMSE series of each CC2 variant and that of CC2 is very short, implying that NP-Free indeed generates similar RMSE series for similar non-recurrent time series.

TABLE III. THE EUCLIDEAN DISTANCE BETWEEN THE RMSE SERIES OF EACH B3B VARIANT AND THE RMSE SERIES OF B3B.

| Variant Name | Euclidean Distance |
|---|---|
| B3B+100 | 0.060 |
| B3B+200 | 0.62 |
| B3B+300 | 0.060 |
| B3B+400 | 0.060 |
| B3B+500 | 0.62 |
| B3B+600 | 0.62 |
| B3B+700 | 0.62 |
| B3B+800 | 0.62 |
| B3B+900 | 0.62 |
| B3B+1000 | 0.216 |
| B3B+1500 | 0.213 |
| B3B+2000 | 0.213 |

TABLE IV. THE EUCLIDEAN DISTANCE BETWEEN THE RMSE SERIES OF EACH CC2 VARIANT AND THE RMSE SERIES OF CC2.

| Variant Name | Euclidean Distance |
|---|---|
| CC2+100 | 0.007 |
| CC2+200 | 0.016 |
| CC2+300 | 0.016 |
| CC2+400 | 0.016 |
| CC2+500 | 0.017 |
| CC2+600 | 0.017 |
| CC2+700 | 0.017 |
| CC2+800 | 0.017 |
| CC2+900 | 0.017 |
| CC2+1000 | 0.23 |
| CC2+1500 | 0.23 |
| CC2+2000 | 0.227 |

Now we continue to demonstrate that NP-Free can generate similar RMSE series for similar *recurrent* time series. To do it, we followed the same procedure to separately create 12 variants for ADSN and BSMS so that all the variants for ADSN have the same data pattern and shape as ADSN, and that all the variants for BSMS have the same data pattern and shape as BSMS.

Due to the page limit, we only show ADSN and one of its variants (called ADSN+100) in Fig. 5. Apparently, they have exactly the same shape and data pattern. Fig. 6 depicts the RMSE series generated by NP-Free for ADSN and ADSN+100. Clearly, it is hard to distinguish the two RMSE series due to the significant overlapping. Hence, we further calculated the Euclidean Distance between the RMSE series of ADSN and the RMSE series of each ADSN variant. As shown in TABLE V, all distance values are very low.

We also obtained similar results when NP-Free was used to generate RMSE series for BSMS and each BSMS variant. As we can see from TABLE VI, all Euclidean Distance values are very low as well. Based on the above experiment results, we confirm that NP-Free can generate similar RMSE series for similar time series.

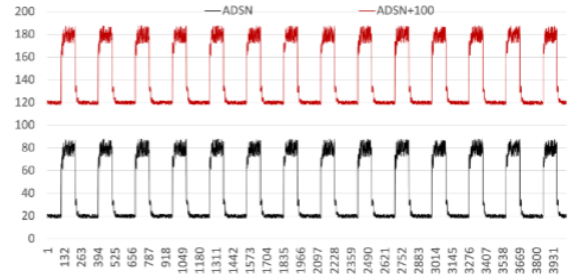

Fig. 5. All data points in ADSN and ADSN+100.

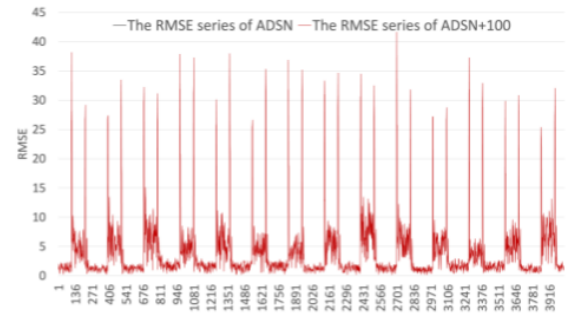

Fig. 6. The RMSE series generated by NP-Free for ADSN and ADSN+100.

TABLE V. THE EUCLIDEAN DISTANCE BETWEEN THE RMSE SERIES OF EACH ADSN VARIANT AND THE RMSE SERIES OF ADSN.

| Variant Name | Euclidean Distance |
|---|---|
| ADSN+100 | 0.08 |
| ADSN+200 | 0.08 |
| ADSN+300 | 0.08 |
| ADSN+400 | 0.08 |
| ADSN+500 | 0.083 |
| ADSN+600 | 0.083 |
| ADSN+700 | 0.083 |
| ADSN+800 | 0.083 |
| ADSN+900 | 0.083 |
| ADSN+1000 | 0.247 |
| ADSN+1500 | 0.247 |
| ADSN+2000 | 0.246 |

TABLE VI. THE EUCLIDEAN DISTANCE BETWEEN THE RMSE SERIES OF EACH BSMS VARIANT AND THE RMSE SERIES OF BSMS.

| Variant Name | Euclidean Distance |
|---|---|
| BSMS+100 | 0.005 |
| BSMS+200 | 0.008 |
| BSMS+300 | 0.011 |
| BSMS+400 | 0.014 |
| BSMS+500 | 0.016 |
| BSMS+600 | 0.016 |
| BSMS+700 | 0.016 |
| BSMS+800 | 0.020 |
| BSMS+900 | 0.022 |
| BSMS+1000 | 0.023 |
| BSMS+1500 | 0.027 |
| BSMS+2000 | 0.035 |

Recall that NP-Free has a lightweight design. To demonstrate that NP-Free is indeed lightweight, we separately evaluated the time consumption of NP-Free in generating RMSE representations for time series B3B, CC2, ADSN, and BSMS. Recall that NP-Free only trains an LSTM model in the beginning and when it finds that current RMSE value is greater than the current value of $thd_{RMSE}$. TABLE VII shows the number of LSTM model training required for each time series. When NP-Free converted B3B into the corresponding RMSE time series, only 59 data points out of 4032 data points triggered LSTM model training, implying that the required training ratio is only 1.46% (=59/4032). It is also clear from TABLE VII that the required ratios for the other three time series are very low, meaning that NP-Free did not need to train LSTM models often.

TABLE VII. TOTAL NUMBER OF REQUIRED LSTM MODEL RETRAINING FOR DIFFERENT TIME SERIES.

| Time Series | Number of data points triggering LSTM model training |
|---|---|
| B3B | 59 out of 4032 |
| CC2 | 61 out of 4032 |
| ADSN | 127 out of 4032 |
| BSNS | 15 out of 4368 |

When LSTM model retraining is required, NP-Free might require more time to generate a RMSE value. TABLE VIII lists the average time required by NP-Free to generate a RMSE value for each data point in the four different time series while LSTM model retraining is required. On the other hand, TABLE IX shows the average time required by NP-Free while LSTM model training is not required. Apparently, when LSTM model retraining is required, NP-Free needs more time to generate a RMSE value. Nevertheless, we can see that NP-Free is very efficient since it does not need to retrain LSTM models very often and that even when it needs to do so, it can generate a RMSE value in a short time. That is why we said NP-Free can generate representations in real time.

TABLE VIII. TIME CONSUMPTION FOR DIFFERENT TIME SERIES WHILE LSTM MODEL RETRAINING IS REQUIRED.

| Time Series | Average Time to Generate a RMSE Value (sec) | Standard Deviation (sec) |
|---|---|---|
| B3B | 0.233 | 0.045 |
| CC2 | 0.241 | 0.061 |
| ADSN | 0.217 | 0.048 |
| BSNS | 0.291 | 0.073 |

TABLE IX. TIME CONSUMPTION FOR DIFFERENT TIME SERIES WHILE LSTM MODEL RETRAINING IS NOT REQUIRED.

| Time Series | Average Time to Generate a RMSE Value (sec) | Standard Deviation (sec) |
|---|---|---|
| B3B | 0.007 | 0.004 |
| CC2 | 0.008 | 0.07 |
| ADSN | 0.008 | 0.003 |
| BSNS | 0.007 | 0.005 |

*B. Experiment Set 2*

The second set of experiments aims to show that NP-Free does not generate similar RMSE series when time series have an opposite data pattern. Recall that RMSE is the square root of the mean of the square of prediction error (see Equation 6). We wonder whether or not the LSTM model employed by NP-Free will generate similar prediction errors for two time series that have an opposite data pattern, and consequently results in two similar RMSE series.

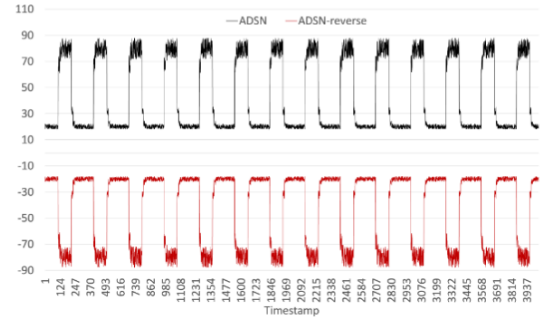

Fig. 7. All data points in ADSN and ADSN-reverse.

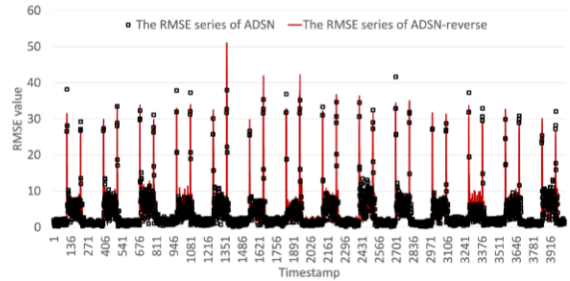

Fig. 8. The RMSE time series generated by NP-Free for ADSN and ADSN-reverse.

To show that the above situation will not happen, we chose ADSN (recall that it is a recurrent time series) and create another variant called ADSN-reverse by reversing every data point in the ADSN time series. In other words, any data point with value 20 in ADSN would become a data point with value -20 in the

ADSN-reverse time series. Fig. 7 illustrates these two time series, and Fig. 8 shows the corresponding RMSE series generated by NP-Free. Apparently, we can see that these two RMSE series are very different from each other, and this is true because their Euclidean Distance is very high (around 94.06).

To further investigate if NP-Free will generate similar RMSE series for time series that have an opposite data pattern, we created one recurrent time series with a sine-wave pattern (see the black one in Fig. 9) and created another time series with an opposite pattern by reversing every data point in the sine-wave time series (see the red one in Fig. 9). Then we employed NP-Free to generate a RMSE series for both series. As Fig. 10 shown, the two RMSE series are very different from each other, and their Euclidean Distance is 32.11. Based on the above experiments, we confirm that NP-Free does not generate similar RMSE series when time series have an opposite data pattern.

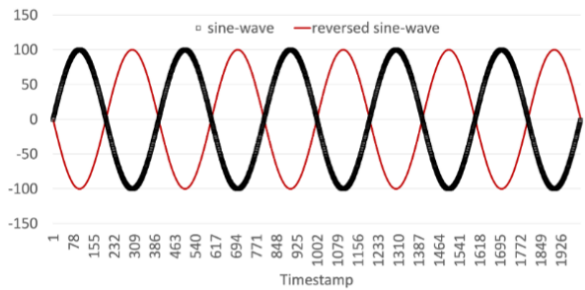

Fig. 9. All data points in the sine-wave time series and the reversed sine-wave time series.

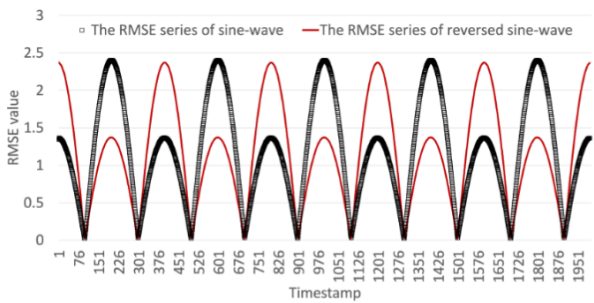

Fig. 10. The RMSE time series generated by NP-Free for the sine-wave time series and the reversed sine-wave time series.

## V. CONCLUSION AND FUTURE WORK

In this paper, we have introduced NP-Free. It can generate a representation for an open-ended univariate time series in real time by keeping predicting each upcoming data point in the time series and calculating the corresponding RMSE value. In other words, NP-Free can convert an open-ended time series into a RMSE series on the fly in real time. Different from existing representation approaches, NP-Free does not need to go through any offline normalization preprocessing. Besides, it does not require users to tune or configure any hyperparameter or parameter.

A series of experiments based on real-world open-source time series datasets have demonstrated that the RMSE series generated by NP-Free can indeed represent the corresponding time series. The results show that NP-Free indeed generates similar RMSE series for similar time series regardless of their offsets in the Y axis, and that NP-Free does not generate similar RMSE series for time series that have an opposite data pattern. Furthermore, the experiments also show that NP-Free is very lightweight because it employs a simple LSTM network (only one hidden layer with ten hidden nodes), always uses three data points to train its LSTM model, and performs LSTM model retraining only when necessary. Therefore, NP-Free can be easily deployed on commodity computers and be used as a processing tool for real-time time series analysis (such as clustering, classification, etc.).

As for future work, we plan to design a real-time clustering algorithm for clustering representations generated by NP-Free. We also would like to propose a real-time large-scale representation clustering approach so that it can cluster large numbers of open-ended representations in real time. This could be helpful for real-time monitoring systems or cyber-physical systems.

## APPENDIX

To demonstrate that z-normalization can destroy potentially relevant properties to distinguish time series, we studied the time series in the GunPointAgeSpan_TRAIN folder of the GunPointAgeSpan dataset[1] of the UEA&UCR archive [14], and found that some time series suffer from the above-mentioned issue.

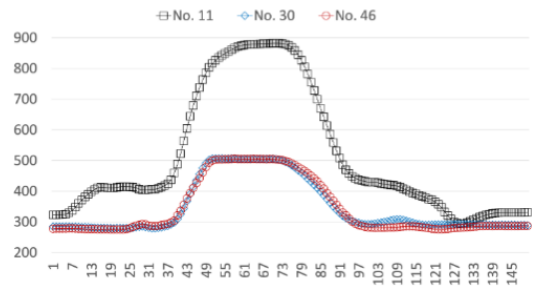

Fig. 11. The data pattern of time series No. 11, No. 30, and No. 46. Apparently, time series No. 11 is different from the other two series.

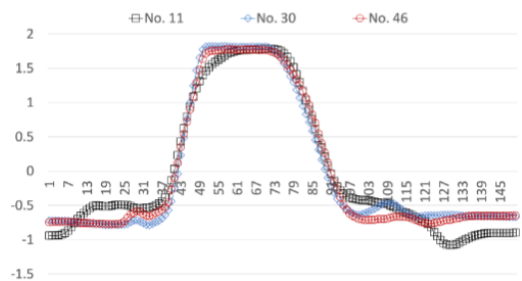

Fig. 12. The z-normalization results of time series No. 11, No. 30, and No. 46. Clearly, they become similar to each other.

Fig. 11 illustrates three raw time series taken from the above-mentioned dataset folder. They are called No. 11, No. 30, and No. 46. It is clear that time series No. 30 and No. 46 are similar to each other, but not series No. 11. However, after z-normalization was applied to these three time series, all of them had a similar pattern as depicted in Fig. 12. If we compare this figure with Fig. 11, we can easily see that z-normalization squeezed time series No. 11. Such an phenomenon might mislead any time series representation approaches that employ z-normalization to generate false representations and consequently negatively affect subsequent data mining tasks. In other words, these three time series might be wrongly clustered into the same group.

---

[1] Please see the following hyperlink for further information.
http://www.timeseriesclassification.com/description.php?Dataset=GunPointAgeSpan.

REFERENCES

[1] S. Hochreiter and J. Schmidhuber, "Long short-term memory," Neural computation, vol. 9, no. 8, pp. 1735--1780, 1997.

[2] M.-C. Lee, J.-C. Lin, and E. G. Gran, "RePAD: real-time proactive anomaly detection for time series," In Proceedings of the 34th International Conference on Advanced Information Networking and Applications (AINA 2020), pp. 1291--1302, 2020. arXiv preprint arXiv:2001.08922 (The updated version of the RePAD algorithm from arXiv was used in this NP-Free paper.)

[3] NAB [Online code repository], Available: https://github.com/numenta/NAB [Online; accessed 12-April-2023]

[4] M.-C. Lee, J.-C. Lin, and E. G. Gran, "How Far Should We Look Back to Achieve Effective Real-Time Time-Series Anomaly Detection?, " International Conference on Advanced Information Networking and Applications (AINA 2021), pp. 136--148, 2021. arXiv preprint arXiv:2102.06560

[5] M.-C. Lee, J.-C. Lin, and E. G. Gran, "ReRe: a lightweight real-time ready-to-go anomaly detection approach for time series, " 2020 IEEE 44th Annual Computers, Software, and Applications Conference (COMPSAC), pp. 322--327, 2020. arXiv preprint arXiv:2004.02319

[6] M.-C. Lee, J.-C. Lin, and E. G. Gran, "SALAD: Self-adaptive lightweight anomaly detection for real-time recurrent time series," 2021 IEEE 45th Annual Computers, Software, and Applications Conference (COMPSAC), pp. 344--349, 2021. arXiv preprint arXiv:2104.09968

[7] F. Höppner, "Less is more: similarity of time series under linear transformations," Proceedings of the 2014 SIAM International Conference on Data Mining, pp. 560--568, 2014.

[8] J. Lin, E. Keogh, L. Wei, and S. Lonardi, "Experiencing SAX: a novel symbolic representation of time series," Data Mining and knowledge discovery, vol. 15, no. 2, pp. 107--144, 2007.

[9] E. Keogh, S. Lonardi, and C. A. Ratanamahatana, "Towards parameter-free data mining," Proceedings of the tenth ACM SIGKDD international conference on Knowledge discovery and data mining, pp. 206--215, 2004.

[10] N. Passalis, A. Tefas, J. Kanniainen, M. Gabbouj, and A. Iosifidis, "Deep adaptive input normalization for time series forecasting," IEEE transactions on neural networks and learning systems, vol. 31, no. 9, pp. 3760--3765, 2019.

[11] E. Keogh and S. Kasetty, "On the need for time series data mining benchmarks: a survey and empirical demonstration," Proceedings of the eighth ACM SIGKDD international conference on Knowledge discovery and data mining, pp. 102--111, 2002.

[12] E. Keogh, K. Chakrabarti, M. Pazzani, and S. Mehrotra, "Dimensionality reduction for fast similarity search in large time series databases," Knowledge and information Systems, vol. 3, no. 3, pp. 263--286, 2001.

[13] H. Ding, G. Trajcevski, P. Scheuermann, X. Wang, and E. Keogh, "Querying and mining of time series data: experimental comparison of representations and distance measures," Proceedings of the VLDB Endowment, vol. 1, no. 2, pp. 1542--1552, 2008.

[14] H. A. Dau et al. "The UCR time series archive," IEEE/CAA Journal of Automatica Sinica, vol. 6, no. 6, pp. 1293--1305, 2019.

[15] P. Senin and S. Malinchik, "Sax-vsm: Interpretable time series classification using sax and vector space model," 2013 IEEE 13th international conference on data mining, pp. 1175--1180, 2013.

[16] S. Aghabozorgi, A. S. Shirkhorshidi, and T. Y.Wah, "Time-series clustering--a decade review," Information systems, vol. 53, pp. 16--38, 2015.

[17] D. D. Paepe, D. N. Avendano, and S. V. Hoecke, "Implications of Z-Normalization in the Matrix Profile," International Conference on Pattern Recognition Applications and Methods, pp. 95--118, Springer, Cham, 2019.

[18] C. Ratanamahatana, E. Keogh, A. J. Bagnall, and S. Lonardi, "A novel bit level time series representation with implication of similarity search and clustering," In Pacific-Asia conference on knowledge discovery and data mining, pp. 771--777, Springer, Berlin, Heidelberg, 2005.

[19] A. Bagnall, et al., "A bit level representation for time series data mining with shape based similarity," Data mining and knowledge discovery, vol. 13, no. 1, pp. 11--40, 2006.

[20] P. Senin, "Z-normalization of time series," https://jmotif.github.io/sax-vsm_site/morea/algorithm/znorm.html [Online; accessed 12-April-2023]

[21] Normalization, https://www.codecademy.com/article/normalization [Online; accessed 12-April-2023]

[22] J. Chung, C. Gulcehre, K. Cho, and Y. Bengio, "Empirical evaluation of gated recurrent neural networks on sequence modeling," arXiv preprint arXiv:1412.3555, 2014.

[23] K. Cho, et al., "Learning phrase representations using RNN encoder-decoder for statistical machine translation," arXiv preprint arXiv:1406.1078, 2014.

[24] M. Santini, "Advantages & Disadvantages of k-Means and Hierarchical clustering (Unsupervised Learning)," http://santini.se/teaching/ml/2016/Lect_10/10c_UnsupervisedMethods.pdf [Online; accessed 12-April-2023]

[25] A. Bagnall, et al., "The great time series classification bake off: a review and experimental evaluation of recent algorithmic advances," Data mining and knowledge discovery, vol. 31, no. 3, pp. 606--660, 2017.

[26] H. Ismail Fawaz, et al., "Deep learning for time series classification: a review," Data mining and knowledge discovery, vol. 33, no. 4, pp. 917--963, 2019.

[27] Root-mean-square deviation, https://en.wikipedia.org/w/index.php?title=Root-mean-square_deviation&oldid=1108025020 [Online; accessed 12-April-2023]

[28] X. Wang, et al., "Experimental comparison of representation methods and distance measures for time series data," Data Mining and Knowledge Discovery, vol. 26, no. 2, pp. 275--309, 2013.

[29] S. Hochreiter, "The vanishing gradient problem during learning recurrent neural nets and problem solutions," International Journal of Uncertainty, Fuzziness and Knowledge-Based Systems, vol. 6, no. 2, pp. 107--116, 1998.

[30] Joe H Ward Jr, "Hierarchical grouping to optimize an objective function," Journal of the American statistical association, vol. 58, no. 301, pp. 236--244, 1963.

[31] J. A. Hartigan and M. A. Wong, "Algorithm AS 136: A k-means clustering algorithm," Journal of the royal statistical society. series c (applied statistics), vol. 28, no. 1, pp. 100--108, 1979.

[32] C. Faloutsos, M. Ranganathan, and Y. Manolopoulos, "Fast subsequence matching in time-series databases," Acm Sigmod Record, vol. 23, no. 2, pp. 419--429, 1994.

[33] "City of Melbourne - Pedestrian Foot Traffic." www.pedestrian.melbourne.vic.gov.au [Online; accessed 12-April-2023].

[34] J. Hochenbaum, O. S. Vallis, and A. Kejariwal, "Automatic anomaly detection in the cloud via statistical learning," arXiv preprint arXiv:1704.07706, 2017.

[35] N. Zou, J. Wang, G.L. Chang, J. Paracha, "Application of advanced traffic information systems: field test of a travel-time prediction system with widely spaced detectors," Transportation Research Record, vol. 2129, no. 1, pp. 62–72, 2009.

[36] T. J. Lee, J. Gottschlich, N. Tatbul, E. Metcalf, and S. Zdonik, "Greenhouse: A Zero-Positive Machine Learning System for Time-Series Anomaly Detection," arXiv preprint arXiv:1801.03168, 2018.

[37] DeepLearning4j, https://deeplearning4j.konduit.ai/ [Online; accessed 12-April-2023].